# A Comparative Evaluation of Population-based Optimization Algorithms for Workflow Scheduling in Cloud-Fog Environments

Dineshan Subramoney
*Department of Computer Science*
*University of the Western Cape*
Cape Town, South Africa
dineshansubramoney@gmail.com

Clement N. Nyirenda
*Department of Computer Science*
*University of the Western Cape*
Cape Town, South Africa
cnyirenda@uwc.ac.za

*Abstract*— This work presents a comparative evaluation of four population-based optimization algorithms for workflow scheduling in cloud-fog environments. These algorithms are as follows: Particle Swarm Optimization (PSO), Genetic Algorithm (GA), Differential Evolution (DE) and GA-PSO. This work also provides the motivational groundwork for the weighted sum objective function for the workflow scheduling problem and develops this function based on three objectives: makespan, cost and energy. The recently proposed FogWorkflowSim is used as the simulation environment with the aforementioned objectives serving performance metrics. Results show that hybrid combination of the GA-PSO algorithm exhibits slightly better than the standard algorithms. Future work will include expansion of the workflows used by increasing the number of tasks as well as adding some more workflows. The addition of some more objectives to the weighted objective function will also be pursued.

## I. INTRODUCTION

Cloud computing is a distributed computing paradigm that provides virtual, scalable and dynamic resources on a pay-as-you-use basis [1]. Facilities offered by cloud computing fall under the following categories: Software as a Service (SaaS), Infrastructure as a Service (IaaS), and Platform as a Service (PaaS) [2]. Some of the benefits of cloud computing include cost efficiency, high speed, excellent accessibility, manageability, elasticity, virtualization capabilities and sporadic batch processing. These attributes have made cloud computing the platform of choice for scientists when they are executing computation-intensive, and collaboration intensive scientific algorithms by using scientific workflows [3-5].

A scientific workflow denotes interdependent tasks and computations aimed at achieving some scientific objectives. Workflows are described as a directed acyclic graph (DAG), where the nodes are tasks and the edges denote the task dependencies [6]. These workflows are characterized by complex data and long sessions of distributed computing. They require high computational resources. Scheduling workflows for execution on cloud resources is accompanied by massive computation and communication costs [7]. It involves mapping of the tasks in the workflows to the available virtual machines in the cloud infrastructure.

Literature is replete with works focusing on the use of population-based algorithms such as Genetic Algorithm (GA) and Particle Swarm Optimization (PSO) for scheduling tasks or workflows in the cloud environment [7-12]. A recent comprehensive survey on Quality of Service (QoS) requirements based on PSO scheduling techniques marks out PSO as the most used population-based optimization approach in this field. Prior to that, there was another survey targeting PSO-based scheduling algorithms in cloud computing [13]. The maximization of the execution finishing time, commonly known as makespan, and minimization of the cost are the two major objectives that have been targeted so far. Other QoS requirements such as deadline and budget are incorporated as constraints.

Population-based optimization techniques for workflow scheduling in cloud environments have so far followed two modes. The first group uses Pareto optimality-based approaches to determine a set of Pareto Optimal solutions [7, 14]. These approaches are motivated by the observation that workflow scheduling has conflicting optimization objectives. A set of non-dominated optimal solutions that satisfy the non-commensurate objectives, irrespective of whether the Pareto front is convex or non-convex is determined. The major drawback is, however, that there is a need for a post-optimization process to extract the best compromise solution from the Pareto optimal set, for practical implementations [15]. The second group combines the objective functions into a weighted sum objective, which is optimized by using single objective optimization algorithms such GA and PSO [8, 9, 11]. These approaches do not have the capability to explore the Pareto front sufficiently when it is non-convex for minimization problems. This problem is, however, not there when the Pareto front is convex [16]. Motivated by the observation that Pareto fronts in the workflow scheduling problem are generally convex [7], this paper adopts the weighted sum approach. This approach can generate sufficiently optimal solutions for the workflow scheduling problem, while averting the need for an *a posteriori* process to determine the best compromise solution [16].

So far, many population-based optimization approaches [8, 9, 11, 12] for workflow scheduling have focused on the two-tier architecture, with the cloud servers at the top and the end devices at the low level. WorkflowSim [17] has been the most common simulation technique for these techniques. Very few works [8, 14] have focused on the emerging three-tier framework that incorporates the fog/edge nodes, as a middle processing infrastructure between the cloud servers and the end devices. The fog layer brings the high computational capabilities of the cloud infrastructure close to end devices. This paper focuses on this three-tier network and it will use the recently proposed FogWorkflowSim [18], which includes a fog layer between the end devices and the cloud.

The major contributions of the work proposed in this paper are as follows:

1. The development of a weighted sum objective function that incorporates makespan, cost, and energy consumption: While makespan and cost are widely used in workflow scheduling, energy efficiency is seldomly used [7, 18]. Nevertheless, with the introduction of fog servers, which have relatively lower energy resources than cloud servers, there is a need to include energy consumption in the optimization process. The energy equation used in this work is holistic in the sense that the end devices, the fog servers as well as the cloud servers are all included. Furthermore, in the development of the energy objection there is a distinction between idle times and active times. As expected, the latter leads to more energy consumption.

2. The introduction of the Differential Evolution (DE) algorithm to scientific workflow scheduling: This technique has been recently applied to task allocation in cloud computing [19], but, to the best of knowledge, it is yet to be applied to the problem at hand.

3. The implementation of the hybrid GA-PSO algorithm [11] in the cloud-fog environments: This algorithm exhibited better performance than PSO and GA in cloud environments.

4. The implementation of the aforementioned algorithms in the FogWorkflowSim [18]: A comparative evaluation of GA, PSO, GA-PSO and DE as optimization tools is also conducted with makespan, cost and energy as performance metrics

The rest of this paper is organized as follows. Section II presents the population-based optimization algorithms used in this work. Section III presents workflow scheduling basics and the formalization of the problem. Section IV presents the workflow optimization process while Section V presents the performance evaluation. Section VI concludes the paper.

## II. A Brief Overview on Population Based Optimization Algorithms used in this work

This presents the four population-based optimization algorithms for scheduling scientific workflows that are evaluated in this paper.

### A. Particle Swarm Optimization

Particle swarm optimization (PSO) is a population-based stochastic optimization method created by Kennedy and Eberhart [20]. It is inspired by the social behavior of bird flocking, fish schooling and other animal societies that act collectively. In this technique, a population of individuals, represented as particles, searches a predefined space according to its current position, $X_i^k$ and current velocity $V_i^k$. Each particle's movement is determined by its best known position $pBest_i$, but is also guided toward the best known position $gBest_i$ for the entire swarm. This process leads the swarm to the best position as the number of iterations increases. The particle's velocity and position are defined using

$$V_i^{k+1} = \omega V_i^k + c_1 r_1 (pBest_i - X_i^k)$$
$$+ c_2 r_2 (gBest_i - X_i^k), \quad (1)$$
$$X_i^{k+1} = X_i^k + V_i^k,$$

where $\omega$ is the inertia weight, $r_1$ and $r_2$ are random numbers between (0,1), and $c_1$ and $c_2$ are the acceleration coefficients (cognitive and social coefficients).

### B. Genetic Algorithm

Genetic algorithms [21] are a class of population-based algorithms that start with a population of randomly generated individuals (represented as chromosomes) and are advanced over a number of iterations toward better solutions by applying genetic operators such as *selection, crossover* and *mutation* analogous to the genetic processes occurring in nature. Standard genetic algorithm first encodes the parameters to generate a certain number of individuals to create the initial population. The algorithm uses the fitness function as the criterion to evaluate the performance of each individual. Genetic operators are used in the creation of new generation of individuals. After creating a new generation of the population at each iteration, the algorithm performs a fitness evaluation of the new individuals. The *elitism* procedure is applied in order to generate the new population by merging the initial population and children. After creating a new generation of the population, the algorithm keeps on performing genetic operations in order to generate new offsprings; the fitness functions for each of the offsprings are evaluated; the best individuals are maintained. This process continues until the end condition is met upon which the individual with the best fitness is returned.

### C. GA-PSO Algorithm

The Hybrid GA-PSO algorithm proposed in [11] is evaluated and compared with the other optimization algorithms. In the first stage, the GA is applied to the entire generated population for the first half of the total iterations, to evolve towards the optimal solution from the existing individuals. The PSO is applied to the population for the following half of the iterations, whereby the resulting chromosomes are passed to the PSO algorithm for the second half of the iterations.

### D. Differential Evolution

Differential Evolution (DE) [22] was first proposed by Storn and Price. The basic process of the DE algorithm is similar to the genetic algorithm. The difference is, however, that DE starts with *mutation*, followed by *crossover* and *selection*, while GA operates in the following order: *selection, crossover, mutation* and *elitism*. DE starts from a randomly generated initial population (each individual represented as a vector of real numbers). During the evolution process of DE algorithm, two different individuals, $X_{r1}^g$ and $X_{r2}^g$, are randomly selected and then subtracted to get a differential vector. The differential weight, $F$, is applied to the differential vector and summed with the third randomly selected individual, $X_{r3}^g$, to generate a variation individual for the mutation process, and then the new individual is compared with the corresponding individual in the current population. The crossover operation involves changing the position of real numbers of the variation individual and target individual to generate a trial individual determined by the crossover probability, $CR$. If the new individual's fitness is better than the current one, the new individual will replace the current one in the next generation, otherwise the old one will remain. Through continuous evolution, the strong individuals are retained, and the inferior individuals are eliminated, which guides the search to approximate the optimal solution.

## III. Workflow Scheduling Basics and Objective Function Formalization

This section starts by presenting the workflow scheduling concept. Then, it proceeds to develop the weighted sum objective function for workflow scheduling.

### A. The Concept of Workflows and Problem Formalization

The workflow is generally modelled as a Direct Acyclic Graph (DAG) [7-11]. The DAG is defined by a tuple $G(T,E)$, where $T = \{t_1, t_2, \ldots t_n\}$ denotes the set of tasks, and $E$ is a set of edges, which denotes temporal dependencies or precedence constraints between pairs of tasks in the workflow. An edge is better visualized by using inter-task data, $d_{ij} = <t_i, t_j> \in E$, where $d_{ij}$ refers to the output data of task $t_i$, which serves as the input for task $t_j$. Therefore, the execution of task $t_j$ will only start after the execution of task $t_i$ has executed. Task $t_i$ is the parent task while task $t_j$ is the child task. The very first task to start executing in a graph does not have a parent; it is called an *entry task* $t_{entry}$. At the other extreme, the final task in a graph does not have any child and it is called an *exit task* $t_{exit}$.

The period between starting time of $t_{entry}$ and the completion time of $t_{exit}$ is referred to as makespan. It is a key measure of performance of workflow scheduling algorithms. The other major measure of performance is the *cost*; it incorporates both the computational cost and the cost of transferring tasks and execution results in the network. As already mentioned, these two metrics are conflicting in nature. Pursuit for shorter makespan implies incurring more computational costs. This is because the processing of the tasks will be predominantly done by the expensive cloud servers and the transfer of task data between the end devices and the cloud servers will also require the user to put in more financial resources. The work in [12] presents a wide range of QoS parameters that have to be incorporated in the workflow scheduling algorithms. Many of the parameters listed in [12] are indirectly addressed by makespan and cost. Energy consumption is, however, one key parameter that is very key to any computational infrastructure, not only from the perspective of cost reduction, but also from the environmental perspective. The energy consumption becomes even more necessary, when the fog servers are included in the computational infrastructure.

### B. The Weighted Sum based Workflow Scheduling Objective Function

In our cloud-fog approach, there are $m$ computational resources which are of three types, namely cloud servers, edge servers and end devices. End devices are included in the mix because for some small tasks, transferring them to the fog and cloud servers does not make economic sense, especially now that some of these end devices are being equipped with advanced computational capabilities and have steady access to power in fixed environments such as homes. Next, we will review the mathematical formulations for makespan, cost and energy consumption in the leadup to the presentation of the weighted sum based objective function.

*1) Makespan*: For task $t_i$ in a particular workflow, let $ST_{t_i}$ and $FT_{t_i}$ denote the starting time and the finishing times respectively. Therefore, *makespan MS* can be determined by using

$$MS = \max\{FT_{t_i}, t_i \in T\} - \min\{ST_{t_i}, t_i \in T\} \quad (2)$$

*2) Cost*: This metric is composed of computational costs and communication cost. Computation costs apply for all the three computational resources while communication costs are not included when the tasks are executed on the end device. The computation cost of using computing resource $r$, which may refer to the end device, the fog server or the cloud server in this work, is defined [8] as

$$CE_i^r = pr * (FT_{t_i} - ST_{t_i}), \quad (3)$$

where $pr$ is unit processing cost of using computing resource $r$. On the other hand, for a particular task the communication cost will only refer to the links between the end device and the fog or the cloud server. The communication cost, incurred when transferring the output file of size $d_{ij}$ from the resource that was processing task $i$ to the resource that will process task $j$, is defined [10] by using

$$CC_{ij} = trc_{ij} * d_{ij}, \quad (4)$$

where $trc_{ij}$ is the unit cost of communication from the resource, where task $i$ is mapped, to the resource, where task $j$ is mapped; $trc_{ij} = 0$ when the two tasks are executed on the same resource. The total cost task $TC$ is, therefore, determined by using

$$TC = \sum_{i=1}^{n}\sum_{j=1}^{n} CC_{ij} + \sum_{r=1}^{m}\sum_{i=1}^{n} CE_i^r. \quad (5)$$

*3) Energy Consumption*: Just like in [10], the energy consumption model has the active and idle components, which are labeled as $E_{active}$ and $E_{idle}$ respectively. The former refers to the energy consumed when a particular task is being executed while the latter is the energy dissipated when the resource is idling. The active energy is defined by

$$E_{active} = \sum_{i=1}^{n} \alpha f_i v_i^2 (FT_{t_i} - ST_{t_i}), \quad (6)$$

where $\alpha$ is the constant; $f_i$ and $v_i$ denote the frequency and the supply voltage for the resource on which task $i$ is being executed. During the idle, the resource goes into sleep mode, where the voltage supply level and the relative frequency are at the lowest level. Therefore, the energy consumed during this time is determined by using [10]:

$$E_{idle} = \sum_{j=1}^{m} \sum_{idle_{jk} \in IDLE_{jk}} \alpha f_{min\,i} v_{min\,i}^2 L_{jk}, \quad (7)$$

where $IDLE_{jk}$ is a set of idling slots on resource $j$ while $f_{min\,i}$ and $v_{min\,i}$ refer to the lowest supply voltage and frequency on resource $j$ respectively; $L_{jk}$ is the amount of idling time for $idle_{jk}$. The total energy consumed in the cloud-fog system for the duration of the workflow is

$$TE = E_{active} + E_{idle}. \quad (8)$$

Therefore, weighted sum objective function, that incorporates makespan, cost and energy consumption, is determined by using

$$F(M) = w_1 * MS_{norm} + w_2 * TC_{norm} + w_3 * TE_{norm}, \quad (9)$$

where $M$ is the mapping of the $n$ tasks of a workflow to the $m$ available computing resources located in the cloud, in the fog and at the end devices; $MS_{norm}$ is normalized makespan, while $TC_{norm}$ and $TE_{norm}$ denote the normalized total cost and the normalized total energy respectively; $w_1$, $w_2$ and $w_3$ are the respective weights that determine the contribution of each of them to the overall objective function. Normalization is necessary here in order to ensure that there are no biases in the realized objective function.

## IV. THE WORKFLOW SCHEDULING OPTIMIZATION PROCESS

This section begins by describing how the workflow tasks are mapped to the available resources and how this mapping is used to generate a solution vector. The second part describes the optimization process based on GA, PSO, DE and GA-PSO.

### A. Mapping of Workflow tasks to computational resources and generation of the solution vector

As already mentioned, the workflows in this work can be scheduled for execution at the source end device, at the fog server or at the cloud server. Each of these computation resources has its own computational power and access bandwidth with respect to the end node. The end nodes do not offload their tasks to fellow end nodes; they can only offload their tasks to fog and cloud servers. Therefore, in the scheduling of workflows, only one representative end device for each of the end nodes is incorporated in the encoding process.

Given that task scheduling in cloud-fog computing environments is a discrete problem, natural numbers are used to encode the individuals for each populated-based algorithm. In the case of the GA, PSO and DE, the individuals represented by the chromosome, particle and agent respectively, are mapped to possible task-resource schedules. The length of each individual is $n$, which is the total number of tasks in the workflow; each position in the individual's vector is a positive integer representing the task number. The value assigned to this position is the virtual machine ID that is allocated to execute the task. The ID numbers are selected from the virtual machines available on the three layers of the cloud-fog architecture. Suppose a workflow has 10 tasks which are scheduled for execution on 5 available virtual machines, specifically one end device, two fog nodes and two cloud servers. In this instance, the individual's length is 10 and each element is an integer between 1 and 5. An example task assignment of this individual can be expressed as {4,3,2,4,5,4,2,1,5,1}. A more detailed representation of the individual's schedule is illustrated in Table I and Table II.

TABLE I. EXAMPLE OF THE INDIVIDUAL'S ENCODED SCHEDULE

| Task number | 1 | 2 | 3 | 4 | 5 | 6 | 7 | 8 | 9 | 10 |
|---|---|---|---|---|---|---|---|---|---|---|
| VM ID | 4 | 3 | 2 | 4 | 5 | 4 | 2 | 1 | 5 | 1 |

TABLE II. EXAMPLE OF THE TASK-RESOURCE ALLOCATION ON THE CLOUD-FOG LAYERS

| VM Layer | Cloud | Cloud | Fog | Fog | End |
|---|---|---|---|---|---|
| VM ID | 1 | 2 | 3 | 4 | 5 |
| Assigned Task | 8,10 | 3,7 | 2 | 1,4 | 5,9 |

### B. The Optimization Process

We will now describe how the four optimization algorithms are implemented. For purposes of brevity, the candidate solutions are simply called vectors for all algorithms instead of the algorithm specific terms such as a particle for PSO and chromosome for GA. Fig. 1 shows the generic flow chart for a population based scheduling algorithm.

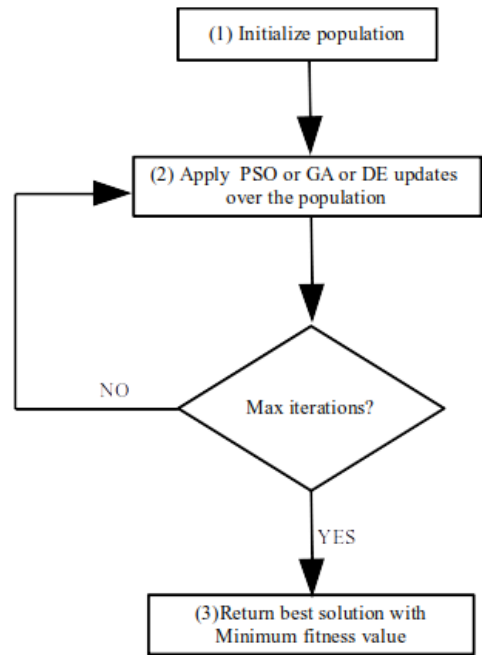

Fig. 1. Flowchart of a generic population-based algorithm for workflow scheduling

In the initialization step, a population of $P$ individuals of length $n$ is initialized with random integer values depicting the possible VM unto which the corresponding task would be assigned. Then workflow simulation for each individual is carried out; this is followed by the determination of the three performance metrics, namely: *makespan MS*, *cost TC* and *energy TE* as described in Section III. The minimum and maximum values of the metrics ($MS_{min}$, $MS_{max}$, $TC_{min}$, $TC_{max}$, $TE_{min}$ and $TE_{max}$) are determined for this initial population, and based on these values, normalized metrics are calculated for each vector by using

$$MS_{norm} = \frac{MS_i - MS_{min}}{MS_{max} - MS_{min}},$$

$$TC_{norm} = \frac{TC_i - TC_{min}}{TC_{max} - TC_{min}}, \quad (10)$$

$$TE_{norm} = \frac{TE_i - TE_{min}}{TE_{max} - TE_{min}},$$

where $MS_i$, $TC_i$ and $TE_i$ are the *makespan*, *cost* and *energy* values for vector $i \in \{1,\ldots,P\}$. The fitness function $F_i$ for the individual is determined by using the weighted sum objective function in (9). After this, the other post initialization processes are more specific for each algorithm. In the GA approach, the worst and the best solutions for the population are saved. In the PSO approach, $\forall i \in \{1,..,P\}\ pBest_i = F_i$ and $gBest_i = min\{pBest_i\}$. In the DE approach, the best solution for the population is saved.

In the second step, algorithm specific routines are applied to the population as follows.

1. In the GA approach, *selection*, *crossover* and *mutation* operators are applied in order to generate a new pool of potential solutions that are evaluated on a workflow simulation. The fitness values for all chromosomes are determined. The elements of the best chromosome and its fitness value are saved. Elitism is applied to truncate the pool of potential solutions to only the best ones, in preparation for the iteration.

2. In the DE approach, *mutation* and *crossover* are performed. The fitness values for all chromosomes are determined. The elements of the best agent and its fitness value are saved. Some of the well performing offsprings replace some agents in the original solution, thereby creating room for further improvements in the next iteration.

3. In the PSO approach, the particle's positions are updated using the PSO equations in (1). The fitness values for all particles are determined. The elements of the best particle and its fitness value are saved. If the new particle positions give a better fitness value than its *pbest*, the *pbest* is updated to the particle's new position; the fitness value is updated accordingly. If one of the particles' updated *pbest* is better than the *gbest*, the *gbest* is updated accordingly, in preparation for the next iteration.

4. The GA-PSO approach applies GA's update mechanism for the first half of the iteration. The PSO update process is applied until the end.

If the maximum number of iterations has not been reached, the algorithm goes back to the second step. This process keeps on getting repeated for each of these algorithms until the maximum number of iterations is reached. Then the best solution with the lowest fitness value is saved in step 3.

The GA and PSO codes are already incorporated into the FogWorkflowSim Simulator [18]. But we integrated the DE and GA-PSO algorithms as well as the weight sum based objective function proposed in this work.

## V. Performance Evaluation

We compare the algorithms using FogWorkflowSim [18]. The FogWorkflowSim is an extensible toolkit built for automatically evaluating resource and task management strategies in Fog Computing with simulated user-defined workflow applications. Next, we describe the workflow models used in this study.

### A. Description of the workflow models

This study uses three well-known scientific workflow applications, Montage, Epigenomics, and CyberShake, which are described and characterized in detail in [23]. The graphical representation of the workflows is shown in Fig. 2. The Montage workflow, created by NASA/IPAC, represents an astronomy application that creates custom mosaics of the sky using multiple input images. The CyberShake workflow is used by the Southern California Earthquake Center to characterize earthquake hazards threatening a region. The Epigenomics workflow, created by the USC Epigenome Center and the Pegasus Team, is used in the field of bioinformatics to automate the different operations in genome sequence processing. These workflows have been extensively used in practice, therefore we have used these workflows in our evaluations conducted to measure the performance of each optimization algorithm.

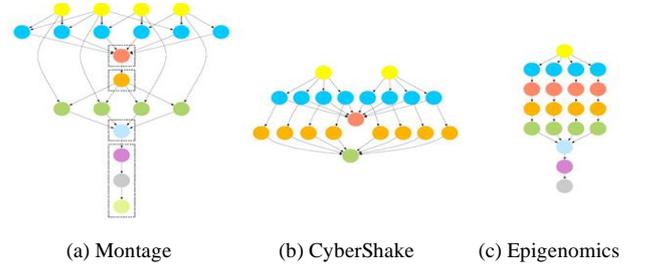

(a) Montage  (b) CyberShake  (c) Epigenomics

Fig. 2. Structure of scientific workflows [23]

### B. Simulation environment

The FogWorkflowSim simulator is run using the Eclipse Java IDE. The simulations are performed on a computer with 64-bit Windows 10 operating system, Intel(R) Core(TM) i5-5200U CPU @ 2.20GHz and 8 GB RAM. The population size was set to 50 for each algorithm. The PSO learning factors C1 = C2 = 2. The inertia weight is 1. The GA crossover and mutation rates are 0.8 and 0.1 respectively. The DE crossover probability is 0.4 and the differential weight is 1.2. The number of iterations for all algorithms is 100. The weighted coefficients $w_1$, $w_2$ and $w_3$ are each set to $0.\bar{3}$. The algorithms are evaluated using the three scientific workflows, with three different task amounts per workflow. Each workflow is a DAG XML file representation of the workflow structures generated by Pegasus [24]. The simulations are performed 10 times for each workflow to get the average performance of the algorithms. Three end devices, 5 fog VMs and 5 cloud VMs are used. The characteristics for each server on the three cloud-fog layers along with the parameter settings for the simulation environment are shown in Table III. The DE used in this code was downloaded from this site [25].

TABLE III. Parameter Settings of Cloud-Fog Computing Environment

| Parameters | End device | Fog node | Cloud server |
|---|---|---|---|
| Processing rate (MIPS) | 1000 | 1300 | 1600 |
| Task execution cost ($) | 0 | 0.48 | 0.96 |
| Communication cost ($) | 0 | 0.01 | 0.02 |
| Working power (MW) | 700 | 800 | 1600 |

| Parameters | End device | Fog node | Cloud server |
|---|---|---|---|
| Idle power (MW) | 30 | 40 | 1300 |
| Uplink bandwidth (Mbps) | 20 | 10 | 1 |
| Downlink bandwidth (Mbps) | 40 | 10 | 10 |

*C. Simulation results*

Fig. 3 - Fig. 5 show the results for makespan, cost and energy consumption for the Montage workflow. As expected, all the metrics increase as the number of tasks increases. Results show that PSO is exhibiting poorer performance compared to the other three approaches. On the other hand, all the three other approaches exhibit similar levels of performance. This is probably because of PSO's well-known problem of premature convergence and getting trapped in the local minimum. Clearly, the GA-PSO algorithm, which adopts the GA approach in the first half of the run and the PSO approach in the latter iterations, seems to benefit from the GA's ability to see a wide range of solutions, courtesy of the random mutation and crossover operators.

Fig. 6 - Fig. 8 show the results for makespan, cost and energy consumption for the Epigenomics workflow. The metrics for 24 and 47 tasks is very low but it springs up when the number of tasks increases to 100. For instance, makespan increases from less than 10 sec for 24 and 47 tasks to beyond 50 sec at 100 tasks; cost increases from less than $10 000 for 24 and 47 tasks to beyond $50 000 at 100 tasks; energy increases from less than 50 000J for 24 and 47 tasks to beyond 250 000J at 100 tasks. This is due to the map jobs in the Epigenomics workflow [23] responsible for aligning sequences with the reference genome, which become significantly more computationally intensive with higher runtimes as the number of tasks increase. In terms of performance, PSO is still exhibiting poorer performance compared to the other three approaches. There is, however, improvement in terms of energy consumption as it performs better than the other approaches for that metric.

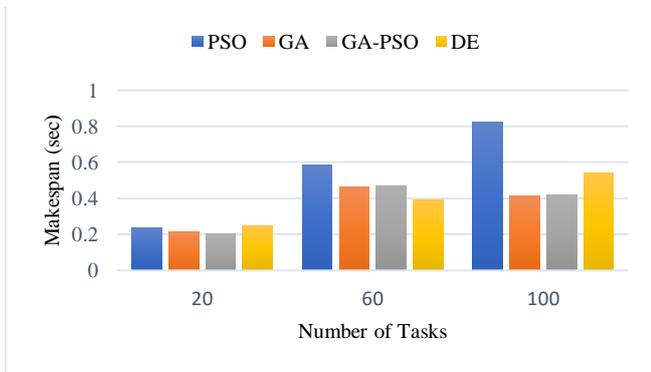

Fig. 3. Makespan for Montage

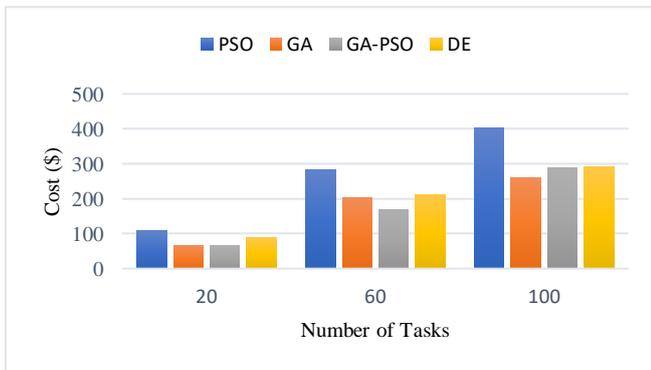

Fig. 4. Total cost for Montage

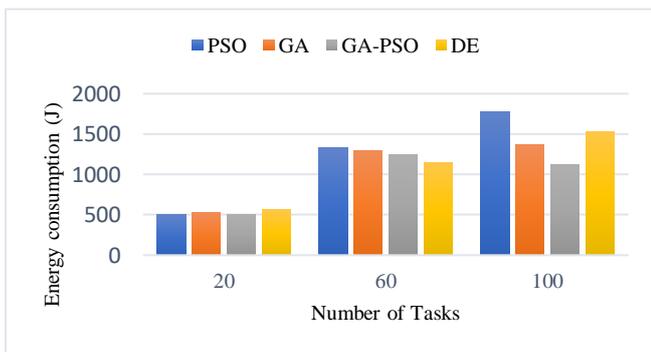

Fig. 5. Energy consumption for Montage

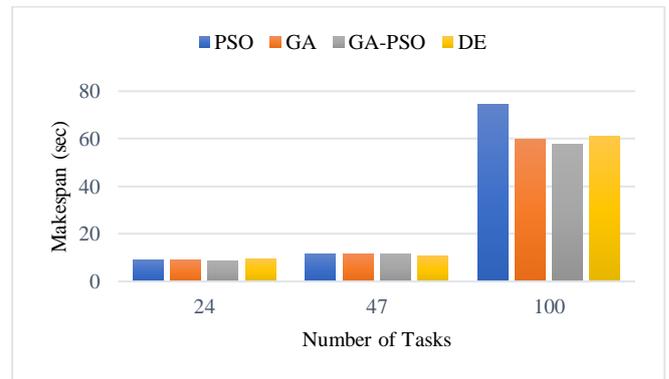

Fig. 6. Makespan for Epigenomics

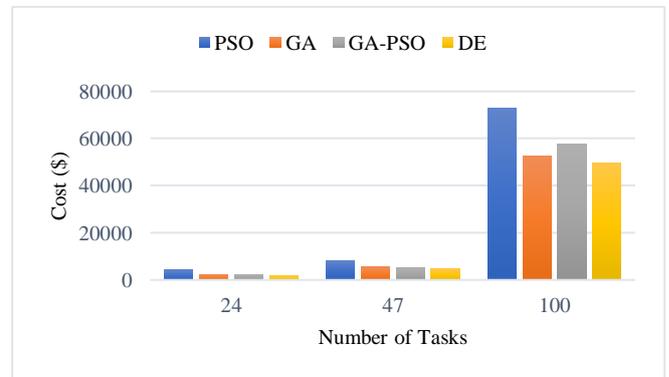

Fig. 7. Total cost for Epigenomics

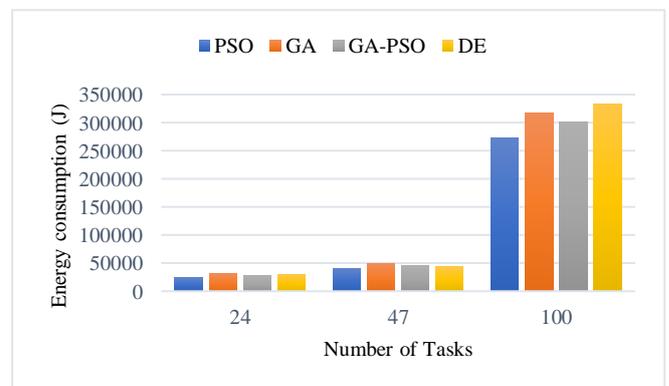

Fig. 8. Energy consumption for Epigenomics

Fig. 9 - Fig. 11 show the results for makespan, cost and energy consumption for the CyberShake workflow. In Fig. 9, the makespan performance remains fairly constant for all algorithms. GA-PSO exhibits the best performance for all tasks while PSO seems to perform better than in the previous two workflows. DE's makespan increases as the number of tasks increases. In Fig. 10, GA-PSO generally continues to perform better than the other approaches for 30 and 50 tasks. DE's cost goes down drastically as the number tasks increases. It looks like the DE algorithm allocates more tasks to the end devices as the number of tasks increases thereby reducing the cost greatly. There are very low data transmission and computation costs due to less usage of cloud and fog servers. This leads to high energy consumption on the end devices as shown by Fig. 11.

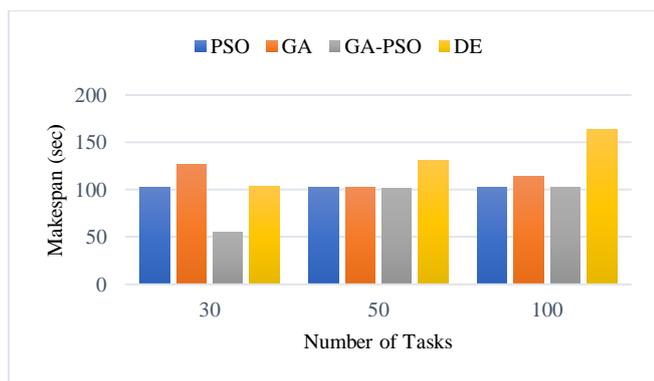

Fig. 9. Makespan for CyberShake

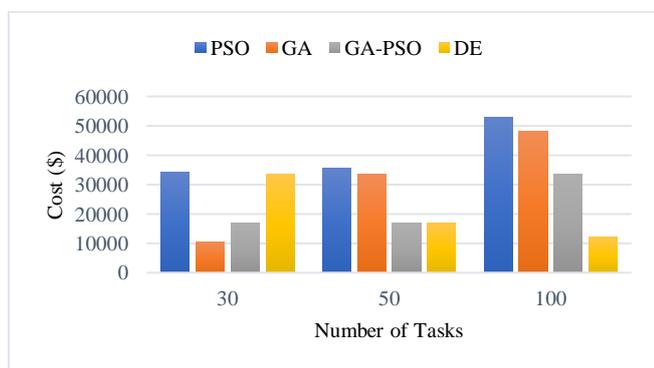

Fig. 10. Total cost for CyberShake

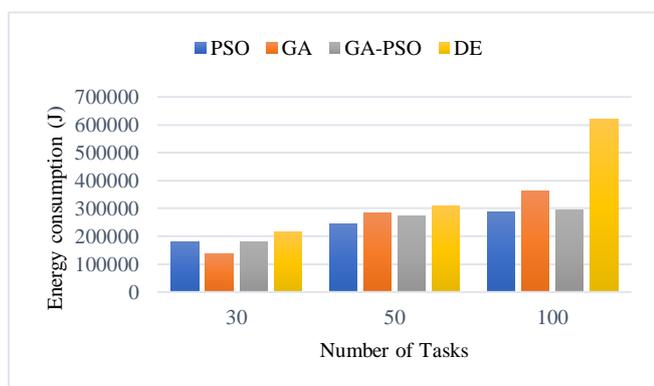

Fig. 11. Energy consumption for CyberShake

The general outcome of these results is that there is no single algorithm that stands out among these algorithms even though the GA-PSO approach seems to exhibit slightly better performance. This shows that there is room for improvement if hybrid algorithms are proposed for the workflow scheduling problem.

In terms of practical implementation, the solution vector realized from the optimization process can be downloaded into the end device as a lookup table for scheduling the workflows.

## VI. CONCLUSION

This work has presented a comparative evaluation of PSO, GA, DE and GA-PSO to the problem of workflow scheduling in cloud-fog environments by using the FogWorkflowSim. It begins by laying out the motivational groundwork in support of the weighted sum method of developing the optimization objective function for workflow scheduling. It proceeds to present the optimization objective by combining makespan, computation and communication costs, and energy consumed in active as well as in idle mode on all the computation devices; this helps to ensure that a holistic view of energy consumption is incorporated in the optimization process. This work has also introduced the Differential Evolution (DE) algorithm to scientific workflow scheduling. It also implements the hybrid GA-PSO algorithm [11] in the cloud-fog environments. Results show that the GA-PSO algorithm exhibits a slightly better performance than the standard approaches. This gives hope for the application of hybrid algorithms that synergistically combine the good attributes of the standard algorithms.

In terms of future work, we intend to further increase the tasks in each of the workflows to 2000. We also intend to add other types of workflows such as LIGO and SIPHT. Plans to expand the number of objectives to include reliability and fault tolerance, and the incorporation of deadline and budget constraints are also in the pipeline. The solution vector realized in this work will be improved by incorporating adjustment mechanisms in the online mode.